\def\year{2020}\relax
\documentclass[letterpaper]{article} 
\usepackage{aaai20}  
\usepackage{times}  
\usepackage{helvet} 
\usepackage{courier}  
\usepackage[hyphens]{url}  
\usepackage{graphicx} 
\usepackage{graphicx}  
\usepackage{color}
\usepackage[]{algorithm2e}
\usepackage{amsmath}
\usepackage{amssymb}
\title{Bias-Aware Heapified Policy for Active Learning}

\author{$^1$Wen-Yen Chang, $^1$Wen-Huan Chiang, $^1$Shao-Hao Lu, $^2$TingFan Wu, $^1$Min Sun \\
$^1$Department of Electrical Engineering,\\
National Tsing Hua University, Taiwan\\
\{s0936100879, wynn8651599, rosyhoward1223\}@gmail.com, sunmin@ee.nthu.edu.tw\\
$^2$Umbo Computer Vision\\
tingfan.wu@umbocv.com\\
}

\begin{document}
\maketitle

\definecolor{gray}{rgb}{0.5,0.5,0.5} 
\definecolor{green}{rgb}{0, 0.4, 0} 
\definecolor{orange}{rgb}{1, 0.5, 0} 	
\definecolor{mahogany}{rgb}{0.75, 0.25, 0.0}
\definecolor{purple}{rgb}{0.6, 0, 0.6}
\definecolor{purple}{rgb}{0.6, 0, 0.6}
\definecolor{darkgreen}{rgb}{0, 0.4, 0} 
\definecolor{frenchblue}{rgb}{0.0, 0.45, 0.73}
\definecolor{unknown}{rgb}{0.5, 0.2, 0}
\definecolor{premod}{rgb}{0.3,0,0}
\definecolor{black}{rgb}{0,0,0}

\newboolean{revising}
\setboolean{revising}{false}
\ifthenelse{\boolean{revising}}
{
	\newcommand{\ignore}[1]{}
	\newcommand{\wy}[1]{\textcolor{frenchblue}{#1}}
	\newcommand{\wyreplace}[2]{\textcolor{frenchblue}{#2}}
	\newcommand{\modify}[1]{\textcolor{mahogany}{#1}}%
	\newcommand{\modifyreplace}[2]{\textcolor{mahogany}{#2}}%
	\newcommand{\wh}[1]{\textcolor{darkgreen}{#1}}
	\newcommand{\whreplace}[2]{\textcolor{darkgreen}{#2}}
    \newcommand{\howard}[1]{\textcolor{unknown}{#1}}
    \newcommand{\howardreplace}[2]{\textcolor{unknown}{#2}}
	\newcommand{\tingfan}[1]{\textcolor{orange}{#1}}
	\newcommand{\tingfanreplace}[2]{\textcolor{orange}{#2}}
	\newcommand{\sunmin}[1]{\textcolor{purple}{#1}}
	\newcommand{\sunminreplace}[2]{\textcolor{purple}{#2}}
	\newcommand{\premodify}[1]{\textcolor{premod}{#1}}
    \newcommand{\premodifyreplace}[2]{\textcolor{premod}{#2}}
    \newcommand{\highlight}[1]{\textcolor{red}{#1}}
} {
	\newcommand{\ignore}[1]{}
	\newcommand{\wy}[1]{#1}
	\newcommand{\wyreplace}[2]{#2}
	\newcommand{\modify}[1]{#1}
	\newcommand{\modifyreplace}[2]{#2}
	\newcommand{\wh}[1]{#1}
	\newcommand{\whreplace}[2]{#2}
    \newcommand{\howard}[1]{#1}
    \newcommand{\howardreplace}[2]{#2}
	\newcommand{\tingfan}[1]{#1}
	\newcommand{\tingfanreplace}[2]{#2}
	\newcommand{\sunmin}[1]{#1}
	\newcommand{\sunminreplace}[2]{#2}
	\newcommand{\premodify}[1]{#1}
	\newcommand{\premodifyreplace}[2]{#2}
	\newcommand{\highlight}[1]{#1}
	
}

\newboolean{format_turning_figure}
\setboolean{format_turning_figure}{true}
\ifthenelse{\boolean{format_turning_figure}}
{
	\newcommand{\FigureFormat}[1]{\vspace*{#1}}
}
{
	\newcommand{\FigureFormat}[1]{}
}
\newboolean{format_turning_Section}
\setboolean{format_turning_Section}{true}
\ifthenelse{\boolean{format_turning_Section}}
{
	\newcommand{\SectionFormat}[1]{\vspace*{#1}}
}
{
	\newcommand{\SectionFormat}[1]{}
}
\newboolean{RLVONE}
\setboolean{RLVONE}{true}

\def \Dl {$D_{l}$} 
\def \Dte {$D_{te}$} 
\def \Deval {$D_{val}$} 
\def \Du {$D_{u}$} 
\def \ActiveModelparam {$\phi\  $}
\def \ActiveModel {$f_{\phi}\ $}
\def \ActiveModelHead {$f_{\hat{\phi}}\ $}
\def \Policy
{$\pi_\theta\ $}
\def \BudgetLimited {$K\ $}
\def \unlabeledsetLimited {$M\ $}
\def \Action
{$A$}
\def \Observation
{$O$}
\def \IspairWOsetn { $\bigcup_{i=0}\{(x_i,y_i)\}$ }
\def \IsnotpairWOsetn {$\bigcup_{i=0}\{x_i\}$} 
\def \etal {\emph{et al. }}

\def \eg {\emph{e.g. }}
\DeclareRobustCommand\bmvaOneDot{\futurelet\@let@token\bmv@onedotaux}
\def\bmv@onedotaux{\ifx\@let@token.\else.\null\fi\xspace}
\newcommand{\EmbeddingFeat}[1] {
  f_{\phi}^{E}({#1})
}
\newcommand{\IspairWsetn}[1] {
  \bigcup_{i=0}^{#1}\{(x_i,y_i)\} 
}
\newcommand{\IsnotpairWsetn}[1] {
  \bigcup_{i=0}^{#1}\{x_i\}
 }
\def \PolicyObjectiveFunction {\nabla\Policy = \max_{\theta} (\log(\Policy(a \mid s;\theta) * reward)
}
\def \Reward {Reward = P_{t+1}-P_{t}}

\def \OurMethod {heapified active learning\ }
\def \OurMethodBrief {HAL\ }
\def \OverConfidenceProblem{overconfidence condition}
\begin{abstract}

\modify{
The data efficiency of learning-based algorithms is more and more important since high-quality and clean data is expensive as well as hard to collect.
}
\wy{
In order to achieve high model performance with the least number of samples, active learning is a technique that queries the most important subset of data from the original dataset.
}
\modify{
In active learning domain, one of the mainstream research is the heuristic uncertainty-based method which is useful for the learning-based system.
}
\wy{
Recently, a few works propose to apply policy reinforcement learning (PRL) for querying important data. It seems more general than heuristic uncertainty-based method}
\wy{
owing that PRL method depends on data feature which is reliable than human prior.
However, there are two problems - sample inefficiency}
\modify{of policy learning} 
\wy{and overconfidence, when applying PRL on active learning. 
To be more precise, sample inefficiency}
\modify{of policy learning}
\wy{occurs when sampling within a large action space, in the meanwhile, class imbalance can lead to the overconfidence.
In this paper, we propose a bias-aware policy network called \OurMethod (\OurMethodBrief \hspace*{-1mm}), which prevents overconfidence, and improves sample efficiency of policy learning by heapified structure without ignoring global inforamtion(overview of the whole unlabeled set).
In our experiment, \OurMethodBrief outperforms other baseline methods on MNIST dataset and duplicated MNIST. Last but not least, we investigate the generalization of the \OurMethodBrief policy learned on MNIST dataset by directly applying it on MNIST-M. We show that the agent can generalize and outperform directly-learned policy under constrained labeled sets.
}

\end{abstract}
\section{Introduction}
\label{Sec:Introduction}
\hspace{3.5mm}
Nowadays, deep learning has been widely used in several fields,
\howard{like medical field or automatic driving.}
However, to reach the whole potential of deep learning, we still struggle to prepare tons of annotated data for training. The progress of collecting such amount of data is obviously a tedious and laborious work.
Thus, it is a critical bottleneck to obtain adequate data for training an accurate model. To solve the problem, simply collecting more data may be the most intuitive idea, yet it is highly time-consuming and expensive in certain domains such as cancer detection, Natural Language Processing tasks, etc.
Thus, active learning comes in handy to minimize the cost by querying important data in order to improve accuracy by labeling as few data as possible.

As to the methodology of active learning, there have been quite a few heuristic methods querying data according to
uncertainty~\cite{Shannon2001Entropy,Jin2014IMSAL,Tang2017ADL},
diversity~\cite{sener2018ICLRCoreset,Wang2017USALDCSS}, etc.
\begin{figure}
    \centering
    \includegraphics[width=\linewidth]{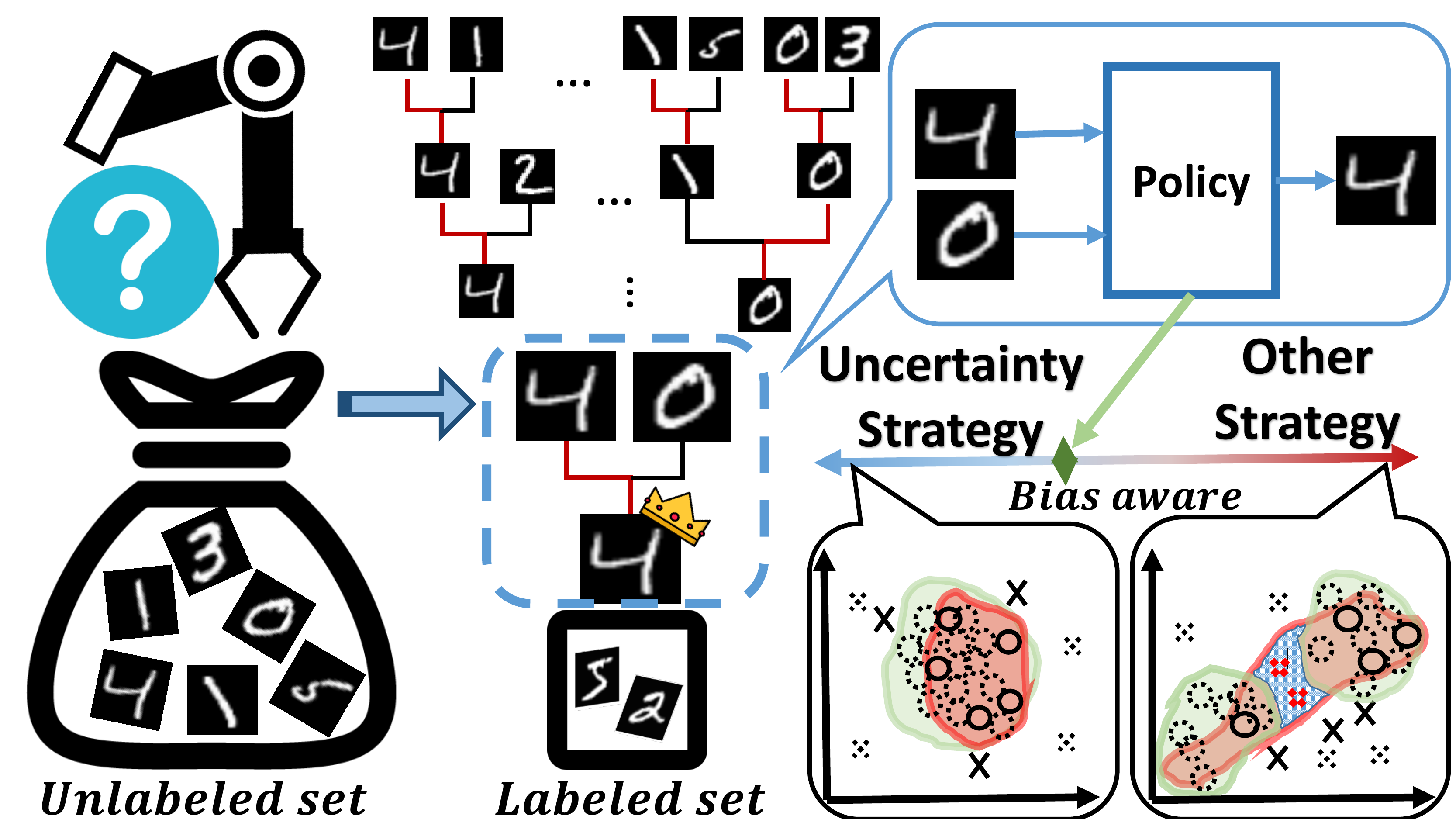}
    \caption{
\small{For selecting the most acceptable image from the unlabeled set, we construct a heap for comparing the whole data two by two which is determined by policy individually. The policy will observe bias-aware feature (Fig.~\ref{fig:BiasAwareConcept}) to avoid using uncertainty-based sampling strategy. After the whole comparison, the final image will be annotated and sent to the training set. This process will be repeated until the budget is exhausted.}
}
\label{fig.teaser}
\end{figure}
Additionally, some work carries out active learning by virtue of multi-heuristic methods such as RALF~\cite{Ebert_RALF_2012_CVPR} which tries to manage different methods with hybrid strategies according to the time, considering that exploration/exploitation criteria should be balanced in different moment.

Nevertheless, \premodify{imbalanced data} is the one that possesses stronger relation to the strategy rather than time, which brings about overconfidence, quite problematic as querying data.
To cope with it, many works
~\cite{sener2018ICLRCoreset,wang2019datasetdistillation,geirhos2018imagenettrained,Olivier2017CoresetConstructions,Balaji2017NIPSUncertaintyEnsembles,Pop2018DEBAL}
intend to eliminate the overconfidence effect. One of the approaches is to query by committee (QBC) \cite{Balaji2017NIPSUncertaintyEnsembles,Pop2018DEBAL}, using the ensemble idea to avoid overconfidence on one single model through multi-models, sacrificing computational cost for training stability.
\wy{Another} approach is to query by disagreement (QBD), directly learning from errors which are probably disagreement samples, including unseen samples and \premodify{in-class potential uncertainty samples, which are predicted incorrectly but possess high confidence from the classification model.}
However, it is a chicken-egg problem that it is hard to identify if the sample is an error before the annotation.
\wy{Thus, some works \cite{gissin2019discriminative,Su_2019_CVPR_Workshops} propose using meta-heuristic method to approximate the idea.
More specifically \cite{gissin2019discriminative} queries data that is likely to be unseen. (unlike the labeled set distribution.)
}
\premodify{However, it is not general enough to model the behavior of in-class potential uncertain samples.
}

\wy{Considering} the generalization of methods, they are not general enough that in every different case, algorithms should be customized. What if we can guide the model with a general agent that shares the experience of querying data?
Recently, there are a few works \cite{Bachman2017LAforALfewshot,Pang2018ICMLMetaActiveLearningbyDeepReinforcement,Ksenia2017LAL,Fang2017Stream-basedRL,Chen_2018_ECCV} focusing on ways of ``learning to active learn''.
Through similar features of data between different datasets, the agent is capable of sharing the experience that is transferred from other datasets.
For example, a stream-based reinforcement learning \cite{Fang2017Stream-basedRL} on prediction of part of the speech, the agent's selection depends on the grammatical architecture of a sentence, which can be applied on both Spanish and Dutch.
Additionally, Chen \etal \cite{Chen_2018_ECCV} propose to take video motion as feature to share experience.
However, in spite of the avoidance of time-consuming by using stream-based active learning, the stability \wy{is} influenced by order of the stream. 

To sum up, the low generalization of current methods for active learning is a critical problem that can result in high cost for designing customized algorithm, and also about how we avoid overconfidence that obstacles uncertainty querying is what we should solve urgently.
In this paper, we propose a bias-aware \OurMethod method for pooling based active learning \wy{called \OurMethod (\OurMethodBrief \hspace*{-1mm}).} It can observe the whole data and spend low computing time with a heapified structure, which is not influenced by the problem of ordering in stream-based reinforcement learning.
Additionally, during querying, we design a bias-aware feature which contributes to the switching of strategy.
Learning with prior of images, the experiences can be transferred to other target domains and remain high performance.
In short, \wy{our \OurMethodBrief \hspace*{-1mm}} enables the agent to make use of features of the training dataset as to query data with the best strategy, avoiding bias which skews the labeled domain. 
In the experiments, our model outperforms other baseline methods on MNIST dataset and synthetic duplicated MNIST dataset which mimics properties in static surveillance videos. Moreover, our policy trained on MNIST achieves the best results on MNIST-M. This demonstrates the great 
\wy{generalization}
of our method across datasets.\\
\textbf{Contribution:}
\begin{itemize}
    \item \wy{ The policy of our methods for querying data is formulated generally which possesses high \wy{generalization} between datasets.}
    \item Our pool based policy agent queries data more efficiently with low time consumption through heapified structure.
    \item Through using bias aware feature, the policy will be able to avoid overconfidence which makes uncertainty based query method inefficiently.
\end{itemize}

In the following sections, we first describe related work in Sec.~\ref{Sec:Related work}.
Then, we introduce our main technical contribution in Sec.~\ref{Sec:Propose Method}, including feature designing, policy setting and so on. 
Finally, we report our experimental results in Sec.~\ref{Sec:Experiment}.

\section{Related Work}
\label{Sec:Related work}
\hspace*{3.5mm}
To improve the efficiency of data sampling, many works tend to use uncertainty-based method and select more sparsing data at the same time in batch mode selection. However, this kind of methods cannot solve the overconfidence problem.
In order to deal with it, in recently, there are three main research directions on active learning related to ours. One is learning with other experts(query by committee, QBC \cite{Pop2018DEBAL,gal2017DBAL,Romer2009QBC1}).
Another is learning from disagreement(query by disagreement, QBD
~\cite{gissin2019discriminative,Kane2017ACCQ,Zhang2014BDisagreementAL,Melanie2018DFAL}).
The other is to manage different criteria and combine the strategies \cite{Pang2018ICMLMetaActiveLearningbyDeepReinforcement,Ksenia2017LAL,Fang2017Stream-basedRL,Chen_2018_ECCV}.
\\ \textbf{Query By Committee.}
The QBC method is to make use of mutual information from multiple models and enable certain sample to be more confident. For instance, DBAL~\cite{gal2017DBAL} generates different outputs through Bayesian network with noises called MC-dropout. However, the outputs generated from different noisy models are not ideal. DEBAL~\cite{Pop2018DEBAL} , following the previous method, proposes the concept of ensembling to solve the problem above. The difference between our method and the ensemble idea is that we use features of labeled data to let the policy model conscious of the situation of overconfidence.\\
\begin{figure*}[t!]
\begin{center}
\includegraphics[width=0.9\linewidth]{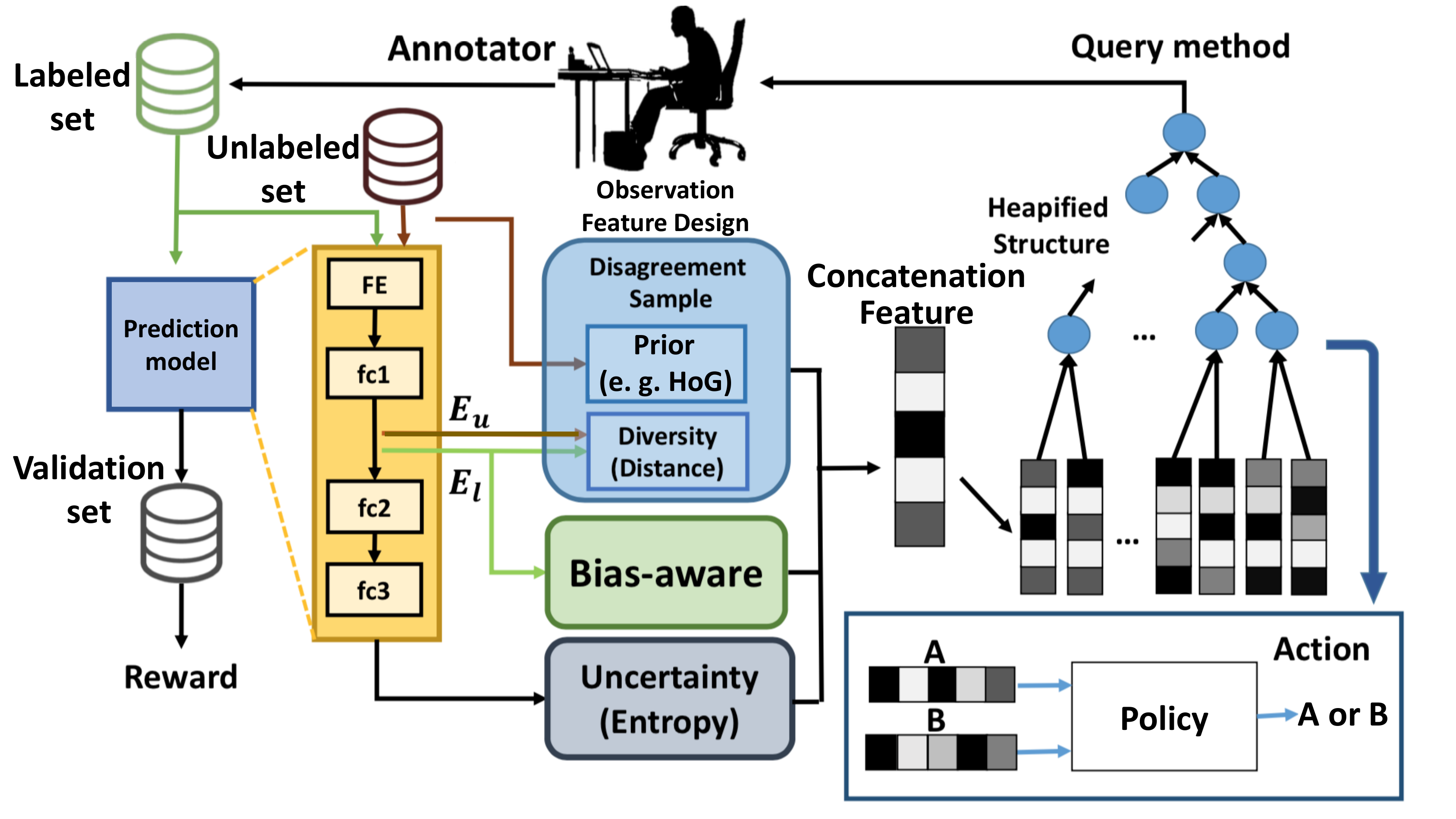}
\caption{
    \small{\wy{Pipeline overview.} First we extract features of every data in \Du \ by $F(x;D_l,\phi)$  and randomly pair all the data together for comparison.($E_l, E_u$ is labeled/unlabeled set embedding feature $\in \mathbb{R}^{f_n}$) In every single comparison with heapified policy, $\pi_{\theta}$, the agent will choose a preferable data. After a series of comparison, the data which is estimated as the most valuable will be annotated and added to the training set \Dl. By training the prediction model \ActiveModel with the new labeled set \Dl \ and evaluating by validation set, \Deval, we can obtain the performance growth of \ActiveModel which can be used as the reward for the agent \Policy.
    }
}
\label{fig:MethodFlow}
\end{center}
\end{figure*}
\hspace*{-1.5mm}\textbf{Query By Disagreement.}
To query by disagreement, we should know which sort of data belong to disagreement cases, thus meta-heuristic exploring strategies are required, like DAL~\cite{gissin2019discriminative} which selects data predicted unlabeled by a discriminator trained with labeled and unlabeled data and DFAL~\cite{Melanie2018DFAL} using adversarial image to build up an image decision boundary, finding its nearby unlabeled sample as uncertain data. However, those meta-heuristic strategies can not be general enough. Our method uses data-driven policy to learn switching strategy for disagreement samples instead, which can be more general in other cases.\\
\textbf{Learning to Active learning.}
Instead of designing the algorithm for selecting unlabeled data heuristically, some adopt stream-based learning \cite{Fang2017Stream-basedRL} by considering it as a decision process with Reinforcement Learning (RL). Through training on Deep Q Network (DQN), it can learn the selecting policy and choose informative data that enable the model to be more robust to certain types of errors. As the number of selected data reaches the budget, the Markov Decision Process (MDP) terminates. 
The state for the agent is composed of the content, marginals and the confidence of prediction. By doing so, the agent can give consideration to both the uncertainty of every word class prediction and the architecture of every sentence which can avoid the bias in the prediction model. At last, taking the performance growth as reward can enable the model to predict the reward of every selection precisely as to make great progress on the performance of the prediction model. However, they trade off time consuming and performance by just reviewing a subset of data and missing more important data in early steps. As for our method, we use pool-based active learning, selecting the most valuable one in the dataset and using heap structure to reduce the time complexity.
\section{Method}
\label{Sec:Propose Method}
\hspace*{3.5mm}
In the following, first, we overview our active learning pipeline in Sec.~\ref{SubSec:OverView}. Second, we describe the design of the observation features for policy learning and \premodify{introduce each of them individually} in Sec.~\ref{SubSec:FeatureDesign}. Third, in Sec.~\ref{SubSec:HeapRL}, \premodify{we propose a new structure of policy which is "heapified" as querying data, and each policy is learned with offline policy gradient.} Before that, we define some common notations below.

Notation : 
We have three sets, labeled set, validation set, and unlabeled set, which are denoted as \{\Dl, \Deval\} = \IspairWOsetn, and \{\Du\}= \IsnotpairWOsetn, where $x_i \in \mathbb{R}^{C\times H \times W}$ is image, we assume there are $L$ classes and denote $y_i \in \{1,2,...,L\}$ as labels.
Besides, we have two models; one is a classification model $f_{\phi}$  with parameter $\phi$, and the other is an agent \Policy with parameter $\theta$. In the classification model, we  extract embedding feature which is denoted as $f^{E}_{\phi}(.) \in \mathbb{R}^{f_n}$.
\subsection{Overview}
\label{SubSec:OverView}
\hspace*{3.5mm} 
As illustrated in Fig.~\ref{fig:MethodFlow}, in our active learning procedure we have a prediction model \ActiveModel supervised by \Dl.
Next, our agent \Policy will repeatedly pick two random samples from \Du \ and compare which \premodify{unlabeled data has more impact on classification model $f_{\phi}$} until the whole \premodify{\Du} has already been compared.
After iterating the comparisons, a final image will be determined and annotated by annotators, and then we add it to \Dl \ for the training of the prediction model \ActiveModel.
Finally, the reward can be calculated by evaluating the marginal performance of the task with the evaluation set \Deval, offering the agent \Policy to learn.
Through the steps mentioned above, the agent \Policy is able to learn a querying policy from the data.
\subsection{Observation Feature Designing} 
\label{SubSec:FeatureDesign}
\hspace*{3.5mm}
As the objective of active learning, we aim to find out the hard samples and the disagreement samples. \premodify{The disagreement samples include unseen samples and the in-class potential uncertain samples. The unseen samples are the data far from labeled set distribution. The in-class potential uncertain samples are predicted incorrectly but possess high confidence from the classification model.} Therefore, feature design can be divided into three parts, bias-aware feature, uncertainty to deal with hard samples, and the disagreement sample learning, respectively.
\begin{figure}[t!] 
\includegraphics[width=\linewidth]{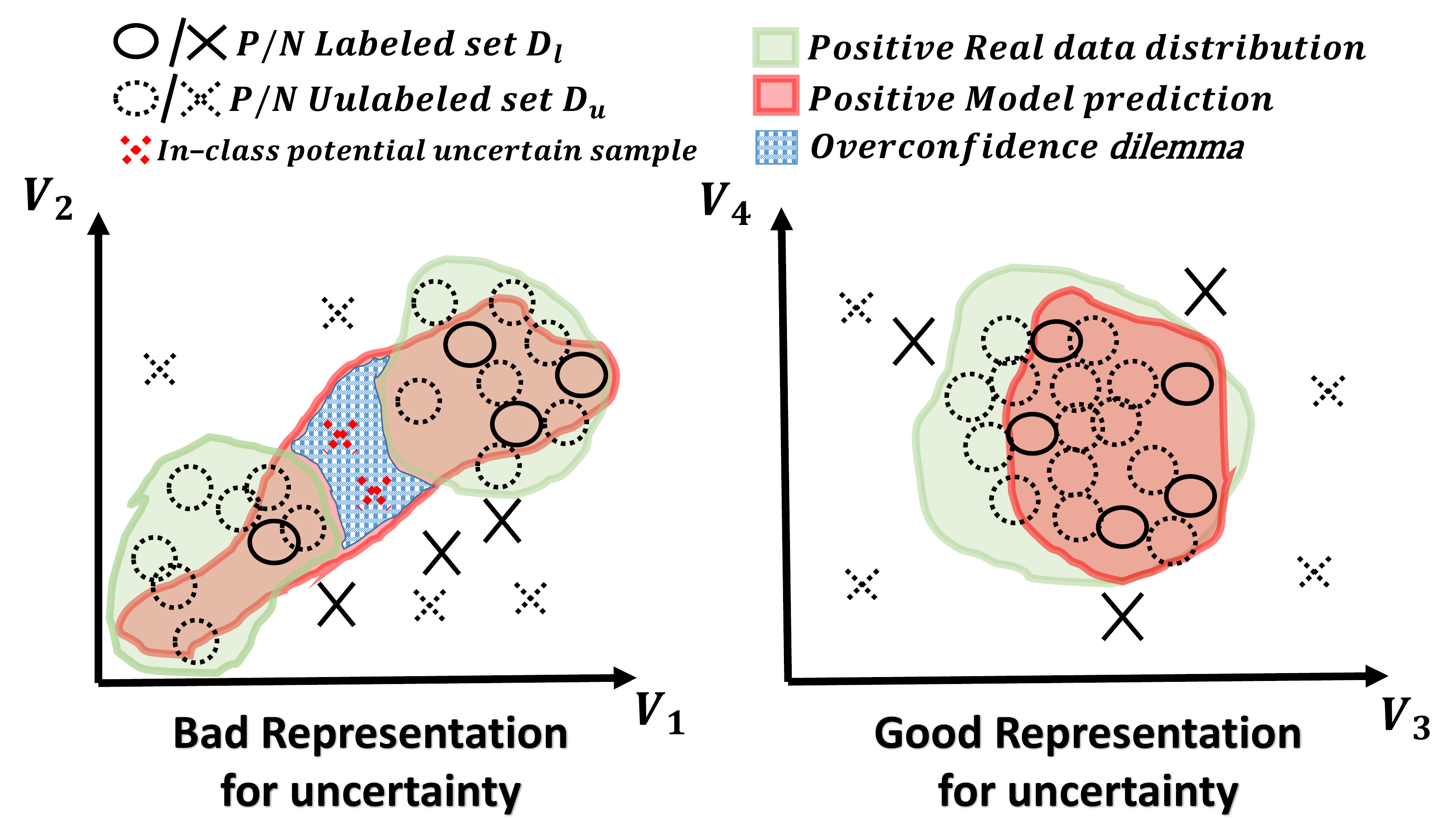}
\caption{
    \small{Bias-aware concept. It's riskier to fall in  \OverConfidenceProblem \  that the eigenspace contribution highly depends on principal eigenvectors, leading to non-continuous distribution of data described by only specific dimensions \premodify{such as the ($V_1$, $V_2$) eigenspace, which have large eigenvalue.}
    Thus, we need to reduce the dependence on principal eigenvectors and get to obtain continuous distribution like the space \premodify{such as the ($V_3$, $V_4$) eigenspace, which have small eigenvalue.}
    }
}
\label{fig:BiasAwareConcept}
\end{figure}
\subsubsection{Bias-aware: maximum component suppression}
\label{SubSec:Biasaware}
\wy{During query procedure, imbalanced data usually results in overconfidence of certain labels, which introduces bias on them.}
 \premodify{Thus, we design bias-aware feature, enabling the agent to observe the distribution of each class for the query policy.
 }
 
 \premodify{
 As shown in  Fig.~\ref{fig:BiasAwareConcept}, overconfidence usually occurs as the real data distribution is non-continuous. Additionally, the blue region in Fig.~\ref{fig:BiasAwareConcept} is where some samples are predicted incorrectly but possess high confidence from the classification model,}
 \premodify{we called that in-class potential uncertainty sample.}
 \premodify{Therefore, in this case, selecting samples with high uncertainty is not an optimal policy. In contrast, if the data distribution is continuous, selecting samples located at the boundary area benefits the training of the classification model.
 }

\premodify{Every data in the dataset can be represented by their own embedding features $f^{E}_{\phi}(.) $ which is extracted from the classification model, indicating that they can be mapped to a multidimensional space.
In each class, through the calculation of the eigenvalues, we can observe the degree of the dominance of each vector.
By taking the largest eigenvalue, the bias-aware feature offers information about the degree of how simply a certain class of data are described. In our design, we select the value oppositely as the feature BA(.) shown in Eq.~\ref{equ:BiasAware},}
\premodify{we called it as maximum component suppression.}
\premodify{
Lower value implies the oversimplification of feature description, causing overconfidence on the unlabeled set \Du.
As a result, the bias-aware feature can be served as a signal enabling the agent to observe the distribution of labeled data \Dl \ for switching different strategies.}
\premodify{Here, we define bias-aware feature as follow:}

\begin{equation}
    BA(\phi,D_l) = 1-\max\lambda_{\EmbeddingFeat{X^{D_l}_{y={y_i}}}} \hspace{0.5em} ,
    \label{equ:BiasAware}
\end{equation}
where $BA(.) \in \mathbb{R^L}$ is the feature of bias aware, $ \lambda_{\EmbeddingFeat{X^{D_l}_{y=y_i}}} $ is the set of class-wise eigenvalues of labeled set's embedding features and this criterion describes how confident can the embedding feature represent the data without main eigenvector.

\subsubsection{Uncertainty}
\label{SubSec:Uncertain}
\premodify{
In order to boost the performance of the classification model \ActiveModel trained on rough data at the beginning of selecting data, we need to find out hard samples located in ambiguous regions near the decision boundary. 
Here, we model it by MC-dropout \cite{gal2017DBAL} which outperforms Shannon entropy. We perturb the model by dropout and compare it with the unperturbed model
so as to find out how uncertain is the data.
That is, the higher the uncertainty of the data, the more it is worth to be selected.
The MC-dropout method is formulated as follow:
}

\begin{equation}
I(x;\phi) \approx H(x;\phi) - \frac{1}{n} \sum_{i=1}^{n} H(x;\phi^{'}_{i}) \hspace{0.5em} ,
\label{equ:MutualInfor}
\end{equation}
\premodify{where the $H(x;\phi) = - \sum_{i=1}^L{P(\hat{y_i}|x;\phi)log(P(\hat{y_i}|x;\phi))}$ , $\hat{y_i}$ is the probability distribution, $\phi$ is the parameters of active model and $\phi^{'}$ is the parameters with noise by dropout which is done $n$ times.
}

However, depending merely on information of uncertainty limits the growth of performance resulting by overconfidence, so we need to solve it by disagreement samples.

\begin{figure*}[t!]
\centering
\includegraphics[width=0.9\linewidth]{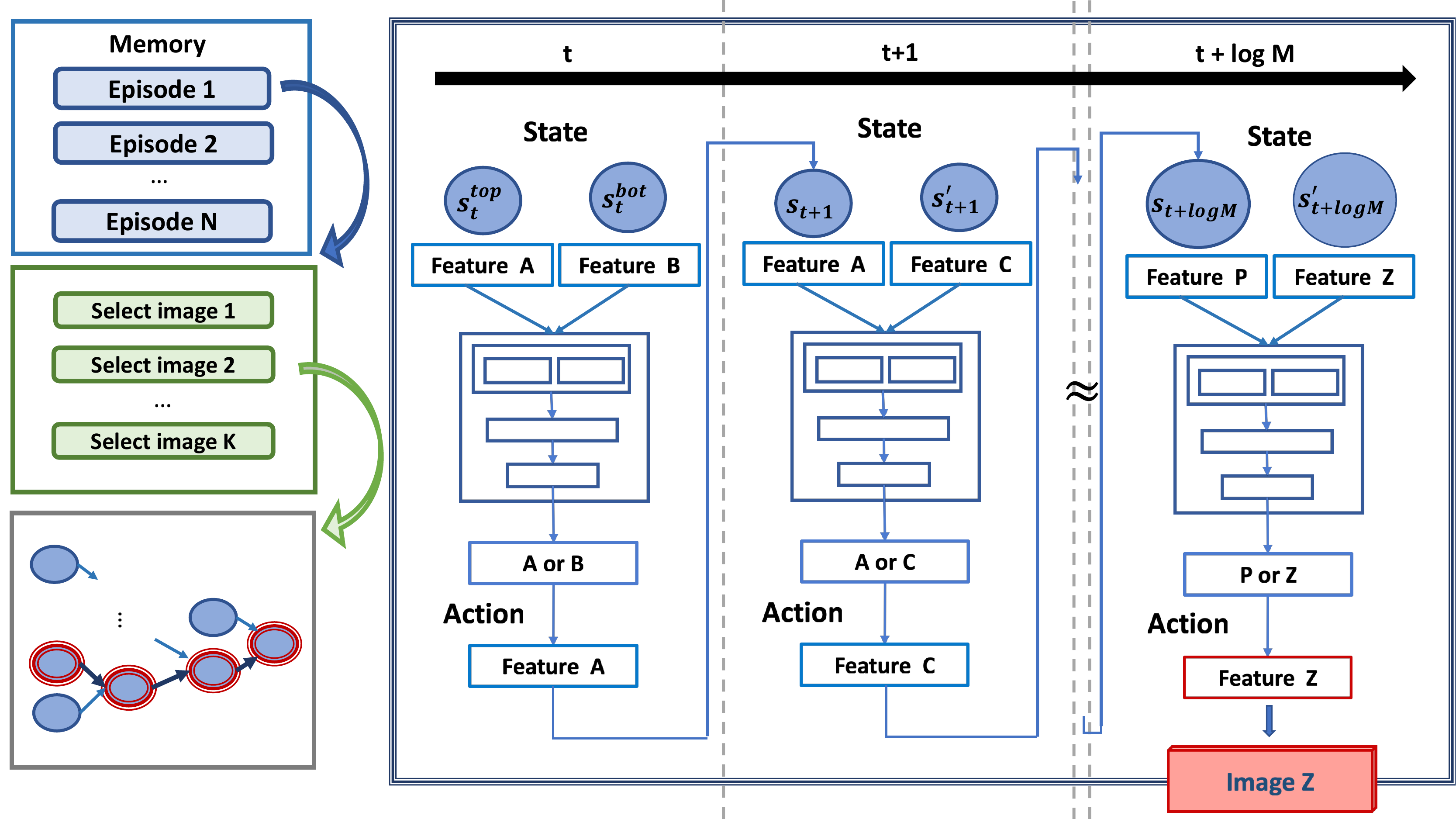}
\caption{
    \small{Off-policy Heapified(compare) policy single selection transition: Here we use memory replay to achieve reward collection efficiently. In every single episode, the agent is required to choose only K images as budget for labeling to the training set \Dl. In each data query, the agent will go through unlabeled set \Du which have M images, and the certain path of the final winner of the whole comparison (dark blue path shown above in the figure) is the most related reward's experience. Policy learns from the path and we show up detail of the decision process with agent network in double box. }
}
\label{fig:RL_agent}
\end{figure*}
\subsubsection{Disagreement sample learning }
\label{SubSec:DisagreementLeaning}
In order to solve the overconfidence samples, we try to use the concept of QBD to learn from disagreement samples \premodify{, which are unseen samples and in-class potential uncertain samples clearly defined in Sec.~\ref{SubSec:Biasaware}}.
To find out unseen sample, inspiring by DAL \cite{gissin2019discriminative}, we query samples that are far from class-wise labeled set distribution. We formulate the calculation of the distance shown as follow:
\begin{equation}
    D(x;\phi,D_l) =\bigcup_{i=1}^{L}Dist(x,\phi,D_{l_{y=i}}) \hspace{1em} ,
    \label{equ:DiversityDist}
\end{equation}
\premodify{where the diversity feature is defined as the distance between unlabeled data and the labeled set representation of each class distribution. The Dist(.) is defined as follow:}
\begin{equation}
    Dist(x,\phi,D_{l_{y=i}}) = norm1({\frac{(\EmbeddingFeat{x}-\bar{\EmbeddingFeat{X^{D_l}_{y={y_i}}}})^2}{2\sigma_{{\EmbeddingFeat{X^{D_l}_{y={y_i}}}}}^2}}) \hspace{1em} ,
    \label{equ:DiversityDistElement}
\end{equation}
where the $x$ is input image sample, \ActiveModelparam is model parameters, $\bar{\EmbeddingFeat{X^{D_l}_{y=y_i}}} \in \mathbb{R}^{f_n}$ is the mean of embedding features in each class of labeled set and ${f_n}$ is the length of the embedding feature.
We have ablation study about labeled set representation in table~\ref{table:AblationStudy}.
\premodify{In the Eq.~\ref{equ:DiversityDistElement}, we calculate the distance between the unlabeled data \Du \ and the labeled data \Dl \ to represent whether data is seen or not for the classification model \ActiveModel.
In addition, we normalize the distance for each class owing that every class distribution is quite different.
}

On the other hand, searching in-class potential uncertain samples for sampling is quite tricky unless we provide handcraft features (\eg SIFT \cite{David2004SIFT}, SURF \cite{Herbert2008SURF}, HOG \cite{Navneet2005HoG}, BoVW \cite{Chandra2012BoVW}) as prior and information of labeled set \Dl \ as constraint.

As designing the feature we observe the model prediction, embedding feature of the unlabeled set \Du \ and some analysis of the labeled set \Dl.The feature of a single data can be represented as follow:
\begin{equation}
    O = \{I(x;\phi),  D(x;\phi,D_l), Prior(x), BA(\phi,D_l)\} ,
    \label{equ:observation}
\end{equation}
using the uncertainty approach of mutual information Eq.~\ref{equ:MutualInfor} as the value of uncertainty for exploiting the mainstream data.
To explore unseen case in labeled set \Dl , we use Eq.~\ref{equ:DiversityDist} to model it. In addition, as to in-class potential uncertain sample which provides a conditional prior, we add a handcraft feature to describe the image statistic information for the active model\ActiveModel to explore more efficiently. Finally, we express the bias-aware feature with \Dl \ by Eq.~\ref{equ:BiasAware} in order to prevent overconfidence which means that in a few classes, misclassification occurs which is caused by low complexity of class features description.

\ifRLVONE
\subsection{Heapified Policy} \label{SubSec:HeapRL}
\hspace*{3.5mm}
Our policy \Policy is a maximum-heap like pooling based query method, so the action space(\Action) is quite large.
Thus, to learn experience more efficiently, we adopt off-policy policy gradient method. As shown in Fig. \ref{fig:RL_agent}\ , our single episode is limited by budget \BudgetLimited and our heapified policy will select the most valuable image from \unlabeledsetLimited unlabeled set images.
Then, we break the task into as many sub-policies, which only compare two features of images, and the better one advances to the next round. 
\premodify{We analyze the time complexity of Monte Carlo experience collection and maximum heapified like sub-policies.
The time complexity of maximum heapified collection is $O(logM)$ less than Monte Carlo collection which is $O(M)$ .
In Monte Carlo sampling method, in order to select the best item, we need to compare pair item $M-1$ times.
On the other hands, maximum heapified collection uses $logM$ times to achieve the goal of the most influential of classification model performance unlabeled data selection.
}
\modify{In this setting, even if the action space is reduced, the global information(overview of the whole unlabeled set) still remains.}
%
\\
\textbf{Sub-Policy model.}
Our sub policy agent $a=\pi_{\theta}(s)$ tries to compare which one is better based on two image's features, \Observation \ define in Eq.~\ref{equ:observation}, where $a \in \{0,1\}$, $ s =(\Observation_1,\Observation_2)$. After two comparisons are done, the two winner data will form the next state, noted as the sub transition $T(s_{t+1}|a^{top}_{t},s^{top}_{t},a^{bot}_{t},s^{bot}_{t})$ as shown in Fig.~\ref{fig:RL_agent}. After we find out the best image, it will merge the sub transitions of the winner into a trajectory. Finally, The agent shall maximize their reward. In our application, we will maximize reward of the marginal accuracy (Acc) of classification task with prediction model $\phi$ as $r = Acc($\Deval$,\phi')-Acc($\Deval$,\phi)$, 
\premodify{where $\phi'$ is trained model parameters and the $\phi$ is original parameters before training.} \\
\textbf{Offline policy gradient.} 
\wy{The reward collection is not efficient and single collection cost much time by reward designed as the increase of performance.}
\wy{Therefore,} we learn from previous sampling reward and decision. Then we compute offline-policy gradient to update model as follow:
\begin{equation}
    \nabla_{\theta}\frac{1}{N}\sum_{j=1}^N\sum_{i=1}^K\sum_{t=1}^{logM}log\pi_{\theta}(a_{i,j,t}|s_{i,j,t})*r*corr  \hspace{1em} .
    \label{equ:PolicyGradientLoss}
\end{equation}
where N is episode of game and the \BudgetLimited is limited of budget. \unlabeledsetLimited is the number of totally unlabeled set images. $\pi_{\theta}(.)$ is now noted as probability estimate.The $corr$ term is to correct the reward which remained from previous policy action probability. The correction term of reward is noted as  $\frac{\pi_{\theta}(a_{i,j}|s_{i,j})}{\pi_{\theta^{'}}(a_{i,j}|s_{i,j})}$, it will maintain the present behavior if identical to previous experience, where $\pi_{\theta^{'}}(.)$ is previous probability estimate, and the $\pi_{\theta}(.)$ is nowstaged probability estimate. The agent will update their gradient direction more stably with previous experience.  

\else
\subsection{Heapified Policy} \label{SubSec:HeapRL}
\hspace*{3.5mm}
\wy{ In pooling-based active learning setting, we need to learn a policy \Policy that the policy select most valuable data by the feature following Sec.~\ref{SubSec:FeatureDesign} definition. 
In this selection, we need to compare whole unlabeled dataset \Du to make the best choice decision.
We use policy gradient (Sec.~\ref{SubSec:PG}) which learn experience by Monte Carlo method due to each comparison is strong correlated in whole comparison process.
The experience is related with model performance improvement in validation set between training before and after (Sec.~\ref{SubSec:Reward design}).
However, in real world setting, experience collection is cost much. 
There have two critical issues need to discuss.
One is the search space too large, we use the maximum-heapfied like learning trick to reduce the length of comparison experience and time complexity from $M$ to $log(M)$ (Sec.~\ref{SubSec:HeapifiedStruture}).
Another is the experience usage is inefficiently by online-policy approach, so we use the offline-policy approach to realize it
(Sec.~\ref{SubSec:Offline-policy policy gradient}).\\
First of all, we introduce the Heapified structure in Sec.~\ref{SubSec:HeapifiedStruture}.
Secondly, we introduce the learning target reward function in Sec.~\ref{SubSec:Reward design}.
Third, we introduce the learning target reward function in Sec.~\ref{SubSec:PG}.
Finally, we introduce the offline-policy correction term in Sec.~\ref{SubSec:Offline-policy policy gradient}.\\
Note : As shown Fig.~\ref{fig:RL_agent}, our single episode is limited by budget \BudgetLimited and our heapified policy will select the most valuable image from \unlabeledsetLimited unlabeled set images.
Then, we break the task into as many sub policies, which only compare two image's features and the better one goes into the next round. The image's features will update after prediction model training.
}
\wy{
\subsubsection{Heapified structure}
\label{SubSec:HeapifiedStruture}
We parallelize the whole comparison trajectory as several same architecture sub policy which share same network in Monte Carlo sampling setting.
Our sub policy agent $a=\pi_{\theta}(s)$ tries to compare which one is better based on two image's features, where $a \in \{0,1\}$, $ s =(\Observation_1,\Observation_2)$ and the learning trajectory only collect the winner path which is most strong related with reward. 
For each pair comparisons are done, the two winner data will form the next state, noted as the sub transition $T(s_{t+1}|a^{top}_{t},s^{top}_{t},a^{bot}_{t},s^{bot}_{t})$ as shown in Fig. \ref{fig:RL_agent}.
After we find out the best image, it will merge the sub transitions of the winner into a trajectory. Finally, The agent shall maximize their reward by Eq.~\ref{equ:reward} define in Sec.~\ref{SubSec:Reward design}.
\subsubsection{Reward design}
\label{SubSec:Reward design}
In our application, we will maximize reward of the marginal accuracy (Acc) of classification task with prediction model $\phi$ as follow:
\begin{equation}
r = Acc(D_{val},\phi)-Acc(D_{val},\phi) ,
\label{equ:reward}
\end{equation}
\subsubsection{Policy gradient}
\label{SubSec:PG}
Policy gradient usually model the long-term status relationship such as data selection in active learning. In active learning, the sequence of data selection will effect model performance and next round data selection. 
Thus, we apply policy gradient to model our policy, the objective formulation as follow:
\begin{equation}
\triangledown J(\theta) = - log(\pi_{\theta}(a,s))\hat{R}
\end{equation}
Where ... .\\
Here, we use policy gradient with baseline prediction in order to learning reward more stable by eliminating variance.
\subsubsection{Offline-policy policy gradient}
\label{SubSec:Offline-policy policy gradient}
Training the policy agent which reward is designed as the increase of performance would cost much. Here we learn from previous sampling reward and decision. Then we compute offline-policy gradient to update model parameters $\theta$ represent as Eq.~\ref{equ:PolicyGradientLoss}.
Where N is episode of game and the \BudgetLimited is limited of budget.
\unlabeledsetLimited is the number of totally unlabeled set images.
$\pi_{\theta}(.)$ is now noted as probability estimate.
The $corr$ term is to correct the reward which remained from previous policy action probability.
The correction term of reward is noted as  $\frac{\pi_{\theta}(a_{i,j}|s_{i,j})}{\pi_{\theta^{'}}(a_{i,j}|s_{i,j})}$, it will maintain the present behavior if identical to previous experience.
Where $\pi_{\theta^{'}}(.)$ is previous probability estimate, and the $\pi_{\theta}(.)$ is nowstaged probability estimate.
The agent will update their gradient direction more stably with previous experience.  
}
\begin{equation}
    \nabla_{\theta}\frac{1}{N}\sum_{j=1}^N\sum_{i=1}^K\sum_{t=1}^{logM}log\pi_{\theta}(a_{i,j,t}|s_{i,j,t})*r*corr  \hspace{1em} .
    \label{equ:PolicyGradientLoss}
\end{equation}
\wy{ 
We want to learn a policy to select most valuable data by design feature as before define. And we need to compare all the data which is the best choice.
During the selection procedure, it's sequence depend with each time selection item.
We use policy gradient to learn the policy, because Monte Carlo method is more reasonable than TD(0) method to learn the trajectory dependency which have long-term relationship.
However, in real world setting, experience collection is cost much. There have two critical issues need to discuss. First, the search space is too large that number of unlabeled data approach infinity, so we divide the comparison Monte Carlo sampling process to several parallel selection process as Heapified like operation.
Second, experience usage is inefficiently in reinforcement learning method, so we apply off-policy method view previous experience as a guiding weight. It's encourage policy to learn previous reward positive if this round experience is similar with previous corresponding experience. 
}
\wy{Pooling-based active learning is a selection process.
We want to learn a policy to build up the best experience selection process by our design feature.
So We apply reinforcement learning approach method to model Markov decision process. 
There are three critical issues need to discuss, one is 
}
\fi
\section{Experiments}
\label{Sec:Experiment}
\hspace*{3.5mm}
We conduct experiments to validate the proposed bias-aware learning to learn policy in cross modalities setting and image duplicated setting. Firstly, in Sec.~\ref{SubSec:AblationStudy} , the result of ablation study shows that our method using bias aware feature and mean representation of labeled set as diversity hint obtain the best result with a few labeled data in the beginning \wy{and the experiment is in train model from scratch setting.}
Secondly, in Sec.~\ref{SubSec:GeneralizationCompare}, we get better result comparing with other baselines in finetune setting.
Finally, we validate the transferability of our query policy across datasets in Sec.~\ref{SubSec:Transferbility}. We report average (15 times) performance of all experiments.

\textbf{Implementation detail.}
We train classification model LeNet5~\cite{Lecun1998Lenet5} with two datasets.
One is MNIST~\cite{lecun2010MnistDatabase}
, and the other one is MNIST-M~\cite{Ganin2015DA_by_backpropagation} which blend background with color photos from BSDS500.
Firstly, we split MNIST, MNIST-M in three subset - labeled, unlabeled and validation set (\Dl, \Du, \Deval), which amounts to (50, 60000, 10000) training pairs with balanced number of class. 
Secondly, we use Adam optimizer~\cite{Kingma2015Adam} with learning rate 0.001 to train our policy agent in 800 episodes. Each episode has 10 steps and each step samples 10 images. The discount factor of policy gradient is set as 0.9998.
Finally, we use the accuracy to plot a learning curve with the size of images and $ALC_{norm}=\frac{ALC-A_{rand}}{A_{max}-A_{rand}}$ which is mentioned in the active learning challenge \cite{Guyon2010ActiveLearningChallengeMetrics}.
Moreover, ALC is the performance of the classification model by proposed query method. $A_{rand}$ is performance of the classification model by random query. $A_{max}$ is performance of the classification model by fully \Du \ with label which will be used in table \ref{table:AblationStudy}.
\ifRLVONE
\begin{table}[t!]
    \centering
    \caption{\small{\wy{Labeled set representation ablation study.} We compare five labeled set representation which are mean, median, mode, minimum and maximum on MNIST with average $ALC_{norm}$.
    \premodify{We find "Mean" is best representation of labeled set in feature space.}}}
    \begin{tabular}{|c||c|c|c|c|c|c|}
    \hline
    \small{Labeled set} & & & & &
    \\ 
    \small{Representation} & 
    \small{Mean} &
    \small{Median} & 
    \small{Mode} & 
    \small{Max} & 
    \small{Min} 
    \\
    \hline \hline
    \small{$mALC_{norm}$} & 
    \small{\textbf{0.207}} &
    \small{0.139} &
    \small{0.148} &
    \small{0.058} &
    \small{0.079}
    \\ \hline
    \end{tabular}
    \label{table:AblationStudy}
\end{table}
\begin{figure}[b!]
    \begin{center}
    \includegraphics[width=\linewidth]{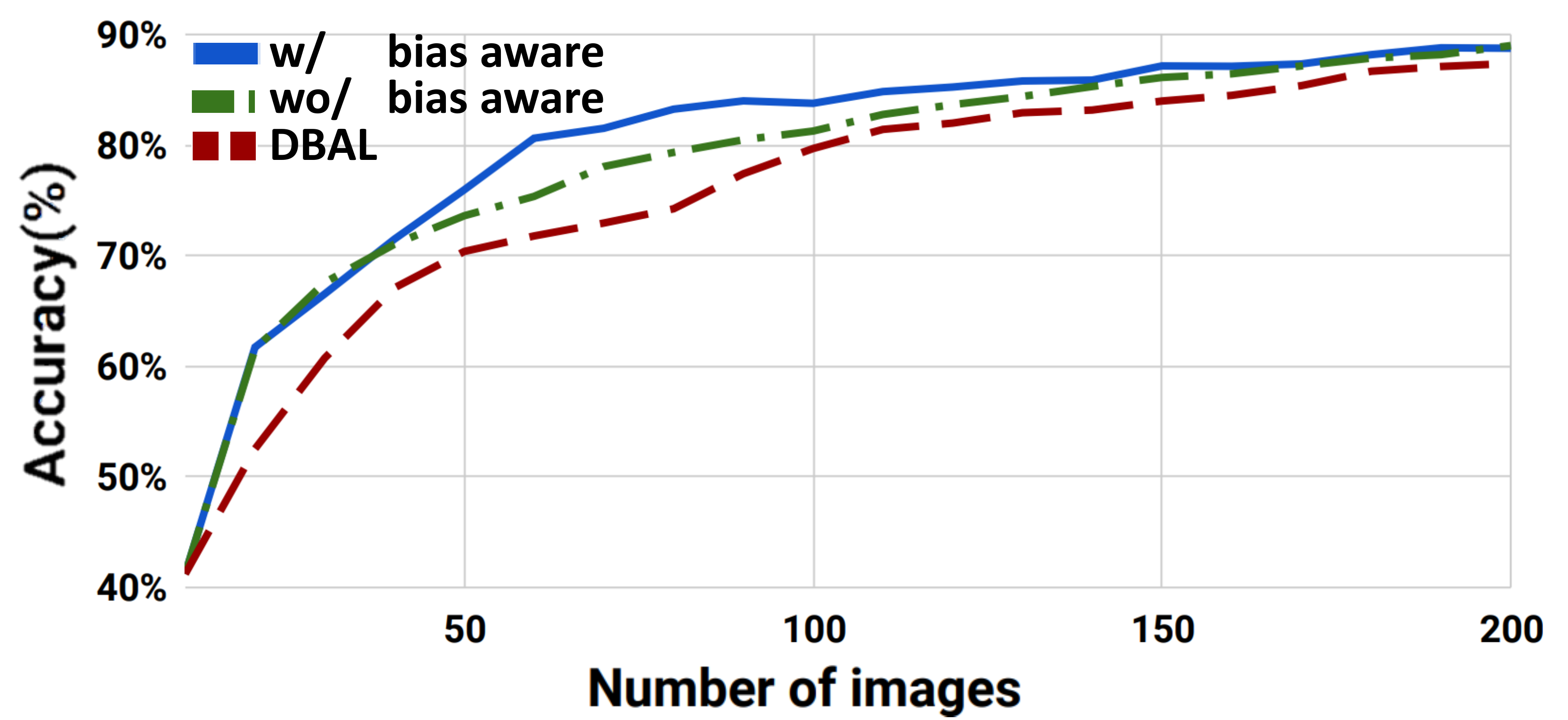}
    \caption{\small{The ablation study of bias aware feature: w/, w/o bias aware comparison. Bias is easily be generated while training from scratch, resulting in uncertainty approach-DBAL will not be useful. Thus, with bias aware feature, overconfidence can be prevented.}
    }
    \label{fig:BiasAwareFeature}
    \end{center}
\end{figure}
\fi
\subsection{Ablation Study}
\label{SubSec:AblationStudy}
\textbf{Diversity feature.}
In the design of the diversity feature, we calculate the distance between the representation of unlabeled set and the labeled set.
\premodify{ There are different kinds of statistic method to represent the diversity feature. We compare these methods, including mean, median, mode, maximum and minimum in table~\ref{table:AblationStudy} by average $ALC_{norm}$ on MNIST.
\ifRLVONE
\else
\begin{table}[b!]
\vspace*{-1em}
    \centering
    \caption{\small{\wy{Labeled set representation ablation study.} We compare five labeled set representation which are mean, median, mode, minimum and maximum on MNIST with average $ALC_{norm}$.
    \premodify{We find "Mean" is best representation of labeled set in feature space.}}}
    \begin{tabular}{|c||c|c|c|c|c|c|}
    \hline
    \small{Labeled set} & & & & &
    \\ 
    \small{Representation} & 
    \small{Mean} &
    \small{Median} & 
    \small{Mode} & 
    \small{Max} & 
    \small{Min} 
    \\
    \hline \hline
    \small{$mALC_{norm}$} & 
    \small{\textbf{0.207}} &
    \small{0.139} &
    \small{0.148} &
    \small{0.058} &
    \small{0.079}
    \\ \hline
    \end{tabular}
    \label{table:AblationStudy}
\end{table}
\fi
As table \ref{table:AblationStudy} shown, we use average $ALC_{norm}$ as an indicator and we can find out \premodify{mean representation} of the class-wise labeled set feature is the best choice.
Thus, we apply mean representation method to represent labeled data and calculate the diversity feature of unlabeled data by Eq.~\ref{equ:DiversityDist}.
}
\ifRLVONE
\begin{figure*}[t!]
\begin{center}
\includegraphics[width=\linewidth]{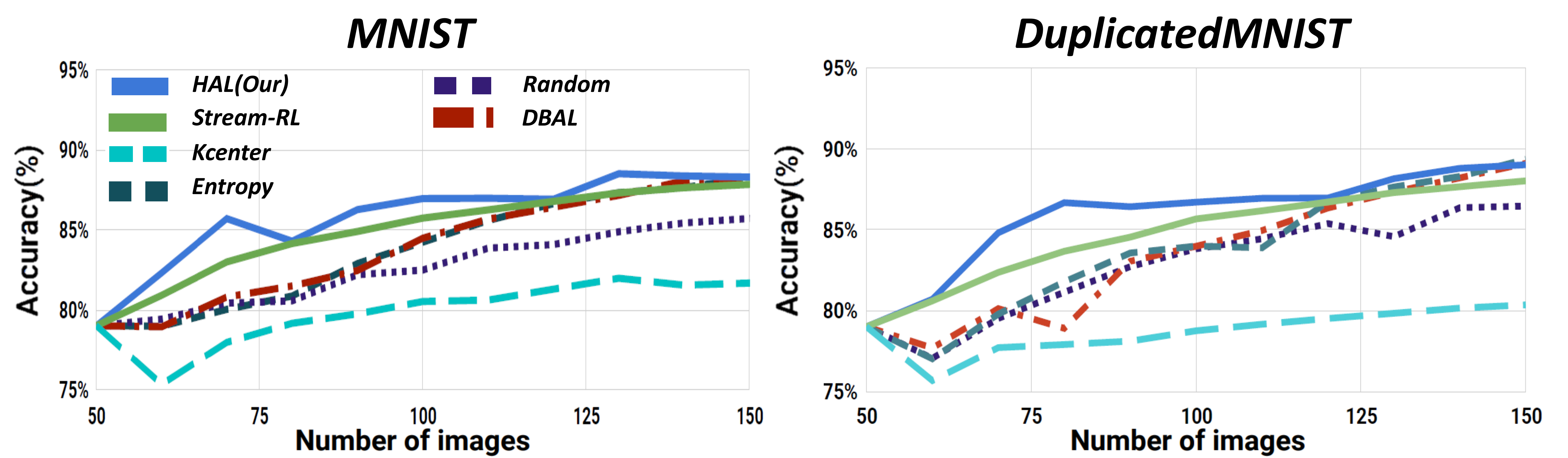}
\caption{\small{Figures above are the average performance of our method and other baselines. On the left figure, we can find out that our \OurMethodBrief only needs less than 100 images to achieve over 85\% of accuracy compared to other methods on average. On the right figure, even in a repeated and noised dataset, \OurMethodBrief can achieve over 85\% of accuracy with less than 75 images; on the contrast, other methods need over 100 images to reach this criterion.}
}
\label{fig:Genralization}
\end{center}
\end{figure*}
\fi
\ifRLVONE
\else
\begin{figure}[t!]
    \begin{center}
    \includegraphics[width=\linewidth]{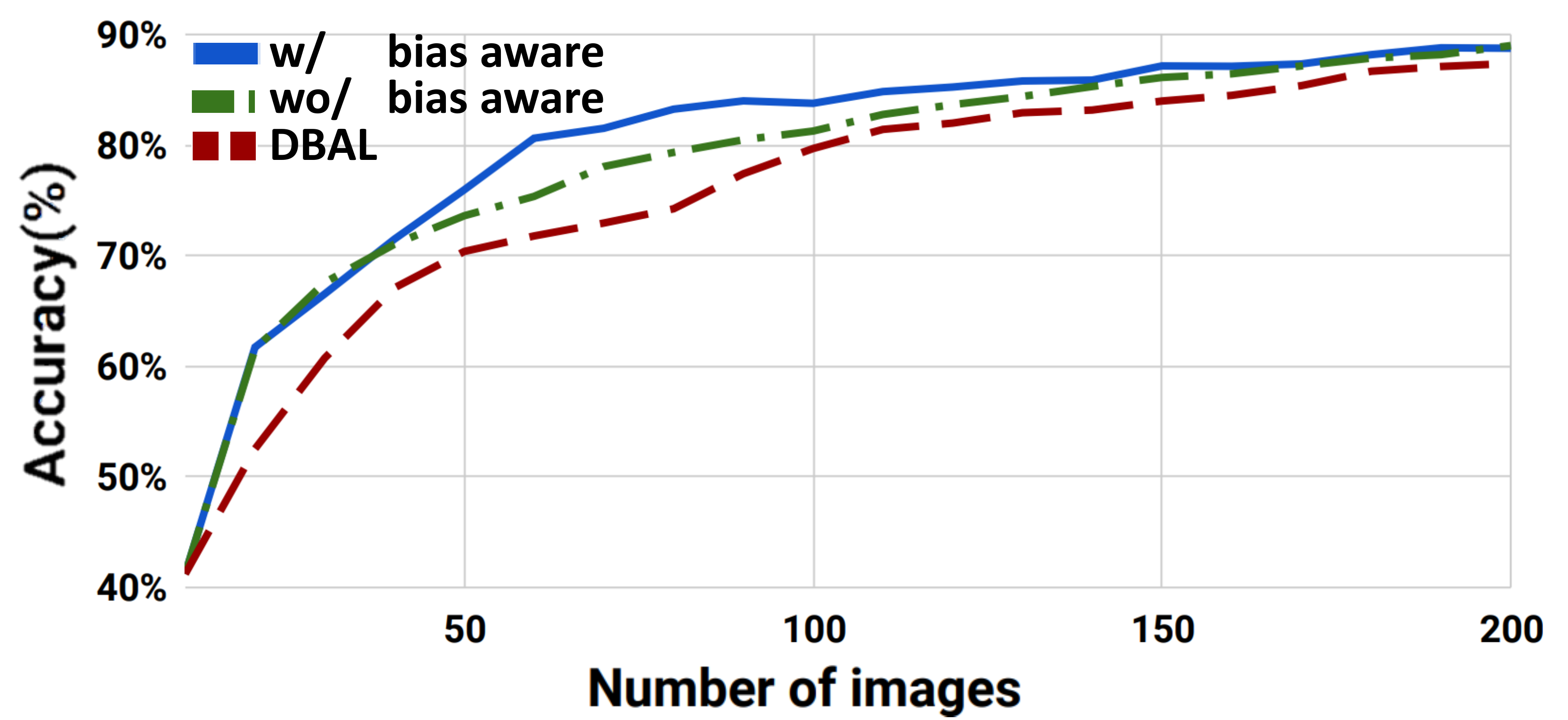}
    \caption{\small{The ablation study of bias aware feature: w/, w/o bias aware comparison. Bias is easily be generated while training from scratch, resulting in uncertainty based method won't be useful. Thus, with bias aware feature, overconfidence can be prevented.}
    }
    \label{fig:BiasAwareFeature}
    \end{center}
\end{figure}
\fi
\textbf{Bias aware feature.}
We design the bias aware feature to avoid \OverConfidenceProblem \  in query data procedure. In order to simulate the dilemma, we train classification model from scratch with little labeled data. Because of \premodify{imbalanced data} (mode collapse~\cite{Pop2018DEBAL} have mentioned), the effect is more extreme on a small dataset, resulting in overconfidence. As Fig.~\ref{fig:BiasAwareFeature} shown, we can see that \OurMethod w/ bias feature can get better performance than w/o bias feature in the interval from 50 to 100 images. 
\modify{The uncertainty approach-DBAL, due to incomplete data understanding, faces overconfidence in the beginning.}
Thus, we know the importance of bias aware feature to avoid overconfidence at the beginning of the query procedure.

\ifRLVONE
\else
\begin{figure}[!b]
\begin{center}
\includegraphics[width=\linewidth]{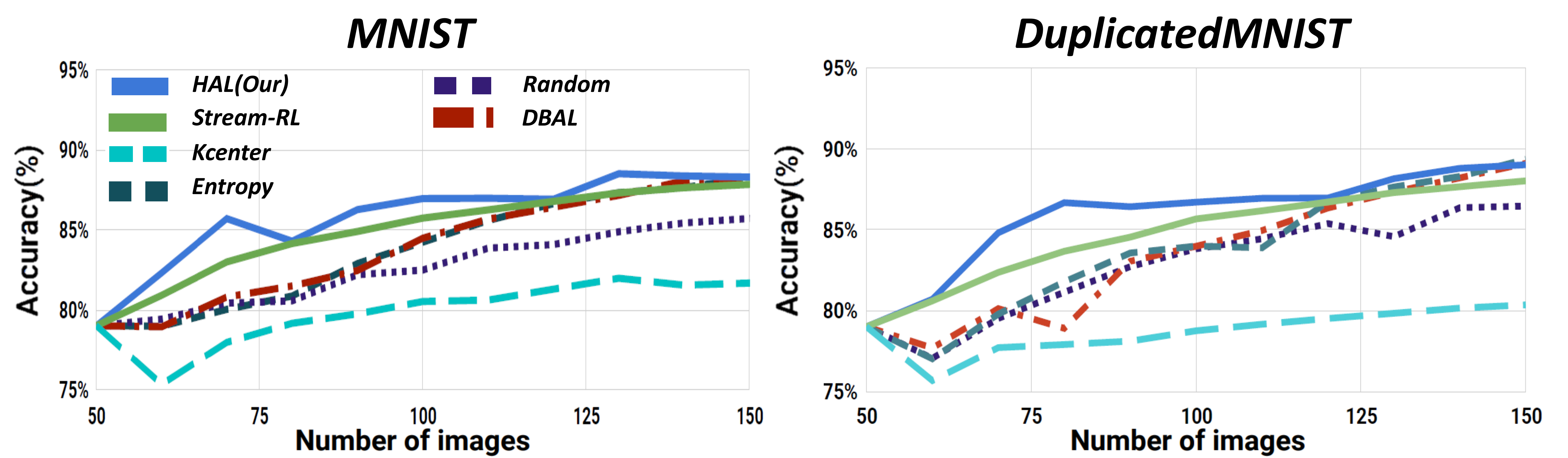}
\vspace*{-1.5em}
\caption{\small{Figures above are the average performance of our method and other baselines. On the left figure, we can find out that our \OurMethodBrief only needs less than 100 images to achieve over 85\% of accuracy compared to other methods on average. On the right figure, even in a repeated and noised dataset, \OurMethodBrief can achieve over 85\% of accuracy with less than 75 images; on the contrast, other methods need over 100 images to reach this criterion.}}
\label{fig:Genralization}
\end{center}
\end{figure}
\fi
\subsection{\wy{Compare Previous Works}}
\label{SubSec:GeneralizationCompare}
\hspace*{3.5mm}
\wy{Here, we compare different types of query methods on MNIST and duplicated MNIST which has many redundant and noise information. From the results, we show that our \OurMethodBrief is outstanding both dataset. Before that, we introduce the baseline methods as following:
}

\begin{itemize}
 \setlength\itemsep{0.01em}
 \item \textbf{Random}: Sample data uniformly from \Du.
 \item \textbf{Entropy}~ \cite{Shannon2001Entropy}: Sample maximum value of chaotic prediction from \Du.
 \item \textbf{DBAL~\cite{gal2017DBAL}}: Apply MC-dropout in the model to produce noises, and then query data with Eq.~\ref{equ:MutualInfor} from \Du \ with maximum value.
 \item \textbf{K-center~\cite{sener2018ICLRCoreset}}: It will compute the minimum Euclidean distance $d$ of an unlabeled data by $kcenter(x_i)=min(x_i,x_j)$, where $x_j \in $\Dl. Then, it will query data with the maximum distance.
 \item \textbf{Stream-based policy network ~\cite{Fang2017Stream-basedRL}}: Through Deep Q Learning (DQN), the agent learns the strategy of choosing images.With the arrival of every batch of images, the agent will decide if the batch of data is necessary to be added to the training set by observing the feature of the batch with the length of action space $len(A)= 2$. As the budget is exhausted, the selection process will be terminated.
\end{itemize}

As shown in Fig.~\ref{fig:Genralization}, our method queries data more efficiently than the other method in the whole training procedure. Instead of the uncertainty based method, they are unstable in the beginning and fall in the  \OverConfidenceProblem.
Specifically, we outperform the stream-based agent on average that it misses many important data in early steps. 
On the other hand, we create a special dataset to test the ability to perform generally among repeated and noised dataset.

\textbf{Synthetic dataset}: In real world application, there may be a lot of redundant data and make the model bias easily.
For example, image data from surveillance camera, it may completely capture the same street view for hours. In this scenario, the capability to avoid duplicate information is essential.
Therefore, we create a synthetic dataset - $Duplicate$ $MNIST$ with 60000 images.
In the set, we have 48000 class-uniformly repeated image (80 percent of the total dataset) with random Gaussian noise.
In the right figure of Fig. \ref{fig:Genralization}, our method is general enough that it is able to achieve high performance in few amounts of data when encountering repeated and noised images. 
\ifRLVONE
\begin{figure}[b!]
    \centering
    \includegraphics[width=\linewidth]{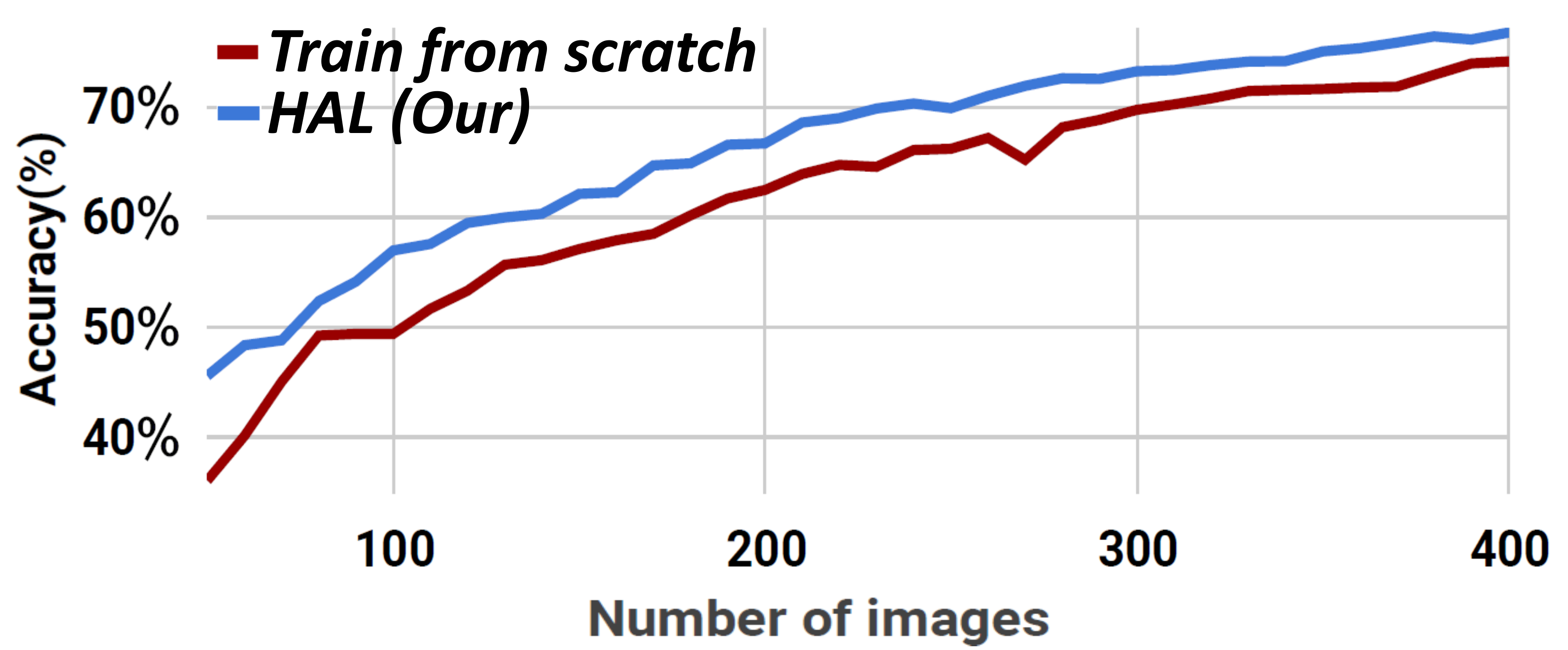}
    \caption{\small{The generalization of query method, \OurMethodBrief. Using our \OurMethodBrief query data directly is more efficiently than training from scratch randomly by  about 5\% performance is Gray(MNIST)$\rightarrow$ RGB(MNIST-M) dataset}
    }
  \label{fig:Transferbility}
\end{figure}
\else
\fi
\subsection{\wy{Generalization}}
\label{SubSec:Transferbility}
\hspace*{3.5mm}
Our method learns how to query from meta-experience with image spatial texture structure as prior by HoG, so we can 
\wy{adopt}
the experience in
cross-domain setting which is Gray(MNIST) v.s. RGB(MNIST-M) scale and outperform with 5\% through querying procedure on average. 
In this setting, we train the agent in MNIST and directly apply as a query method in MNIST-M. 
We obtain a better result than training from scratch randomly shown in Fig.~\ref{fig:Transferbility}.
In this setting, we realize that our \OurMethodBrief is a \wy{general} method can \wy{adopt} query experience to other works that have similar prior.

%
%
\ifRLVONE
\else
\begin{figure}[t!]
    \centering
    \includegraphics[width=\linewidth]{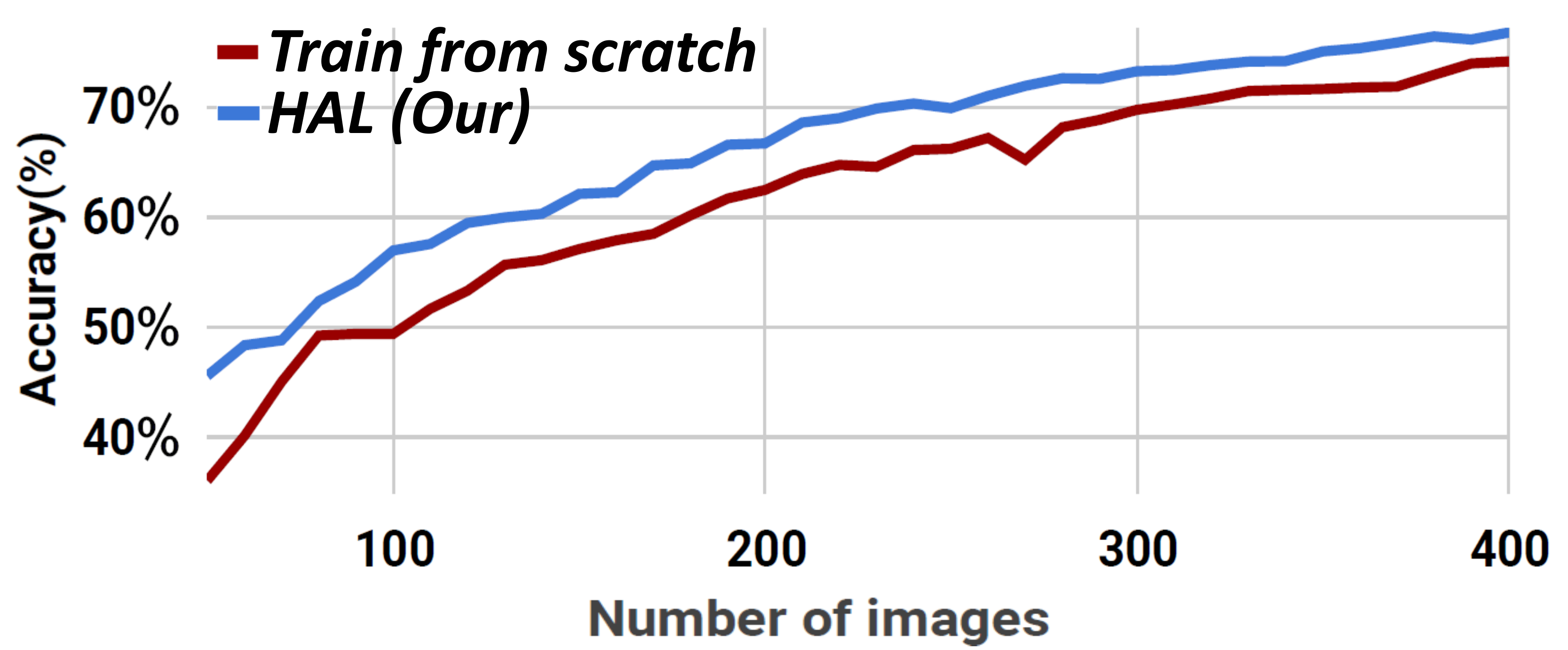}
    \vspace*{-2.5em}
    \caption{\small{The generalization of query method, \OurMethodBrief. Using our \OurMethodBrief query data directly is more efficiently than training from scratch randomly by  about 5\% performance is MNIST$\rightarrow$ MNIST-M dataset}
    }
  \label{fig:Transferbility}
  \vspace*{-0.2em}
\end{figure}
\fi
\section{Conclusion}
\label{Sec:Conclusion}
\wy{
\hspace*{3.5mm}
We proposed a bias-aware policy network called \OurMethod (\OurMethodBrief \hspace*{-1mm}), which prevents data sample bias due to overly confident model prediction.
Moreover, our policy model trades off the query time complexity and global information by heapified structure in pooling based active learning setting.
In addition, in our experiment, \OurMethodBrief outperforms other baseline methods on MNIST dataset and duplicated MNIST. From the results, we can show that our method is able to reach high generalization on different dataset which share similar features.
}

\newpage
\bibliographystyle{aaai}
\bibliography{AAAI}

\begin{thebibliography}{}

\bibitem[\protect\citeauthoryear{Bachem, Lucic, and
  Krause}{2017}]{Olivier2017CoresetConstructions}
Bachem, O.; Lucic, M.; and Krause, A.
\newblock 2017.
\newblock Practical coreset constructions for machine learning.
\newblock {\em ArXiv}.

\bibitem[\protect\citeauthoryear{Bachman, Sordoni, and
  Trischler}{2017}]{Bachman2017LAforALfewshot}
Bachman, P.; Sordoni, A.; and Trischler, A.
\newblock 2017.
\newblock {Learning Algorithms for Active Learning}.
\newblock {\em ArXiv}.

\bibitem[\protect\citeauthoryear{Bay \bgroup et al\mbox.\egroup
  }{2008}]{Herbert2008SURF}
Bay, H.; Ess, A.; Tuytelaars, T.; and Van~Gool, L.
\newblock 2008.
\newblock Speeded-up robust features (surf).
\newblock {\em Comput. Vis. Image Underst.}

\bibitem[\protect\citeauthoryear{Chandra, Kumar, and
  Jawahar}{2012}]{Chandra2012BoVW}
Chandra, S.; Kumar, S.; and Jawahar, C.~V.
\newblock 2012.
\newblock {Learning Hierarchical Bag of Words Using Naive Bayes Clustering}.
\newblock In {\em Asian Conference on Computer Vision}.

\bibitem[\protect\citeauthoryear{Chen \bgroup et al\mbox.\egroup
  }{2018}]{Chen_2018_ECCV}
Chen, Y.-T.; Chang, W.-Y.; Lu, H.-L.; Wu, T.; and Sun, M.
\newblock 2018.
\newblock Leveraging motion priors in videos for improving human segmentation.
\newblock In {\em The European Conference on Computer Vision}.

\bibitem[\protect\citeauthoryear{Dalal and Triggs}{2005}]{Navneet2005HoG}
Dalal, N., and Triggs, B.
\newblock 2005.
\newblock Histograms of oriented gradients for human detection.
\newblock In {\em Proceedings of the 2005 IEEE Computer Society Conference on
  Computer Vision and Pattern Recognition (CVPR'05)}.

\bibitem[\protect\citeauthoryear{Ducoffe and Precioso}{2018}]{Melanie2018DFAL}
Ducoffe, M., and Precioso, F.
\newblock 2018.
\newblock Adversarial active learning for deep networks: a margin based
  approach.
\newblock {\em ArXiv}.

\bibitem[\protect\citeauthoryear{Ebert, Fritz, and
  Schiele}{2012}]{Ebert_RALF_2012_CVPR}
Ebert, S.; Fritz, M.; and Schiele, B.
\newblock 2012.
\newblock {RALF:} {A} reinforced active learning formulation for object class
  recognition.
\newblock In {\em In: IEEE Conf. on Computer Vision and Pattern Recognition}.

\bibitem[\protect\citeauthoryear{Fang, Li, and
  Cohn}{2017}]{Fang2017Stream-basedRL}
Fang, M.; Li, Y.; and Cohn, T.
\newblock 2017.
\newblock Learning how to active learn: A deep reinforcement learning approach.
\newblock In {\em Proceedings of the 2017 Conference on Empirical Methods in
  Natural Language Processing}.

\bibitem[\protect\citeauthoryear{Gal, Islam, and
  Ghahramani}{2017}]{gal2017DBAL}
Gal, Y.; Islam, R.; and Ghahramani, Z.
\newblock 2017.
\newblock Deep bayesian active learning with image data.
\newblock In {\em Proceedings of the 34th International Conference on Machine
  Learning}.

\bibitem[\protect\citeauthoryear{Ganin and
  Lempitsky}{2015}]{Ganin2015DA_by_backpropagation}
Ganin, Y., and Lempitsky, V.
\newblock 2015.
\newblock Unsupervised domain adaptation by backpropagation.
\newblock In {\em Proceedings of the 32nd International Conference on Machine
  Learning}.

\bibitem[\protect\citeauthoryear{Geirhos \bgroup et al\mbox.\egroup
  }{2019}]{geirhos2018imagenettrained}
Geirhos, R.; Rubisch, P.; Michaelis, C.; Bethge, M.; Wichmann, F.~A.; and
  Brendel, W.
\newblock 2019.
\newblock Imagenet-trained {CNN}s are biased towards texture; increasing shape
  bias improves accuracy and robustness.
\newblock In {\em International Conference on Learning Representations}.

\bibitem[\protect\citeauthoryear{Gissin and
  Shalev-Shwartz}{2019}]{gissin2019discriminative}
Gissin, D., and Shalev-Shwartz, S.
\newblock 2019.
\newblock Discriminative active learning.

\bibitem[\protect\citeauthoryear{Guyon \bgroup et al\mbox.\egroup
  }{2011}]{Guyon2010ActiveLearningChallengeMetrics}
Guyon, I.; Cawley, G.~C.; Dror, G.; and Lemaire, V.
\newblock 2011.
\newblock Results of the active learning challenge.
\newblock In {\em Active Learning and Experimental Design workshop In
  conjunction with AISTATS 2010}.

\bibitem[\protect\citeauthoryear{{Kane} \bgroup et al\mbox.\egroup
  }{2017}]{Kane2017ACCQ}
{Kane}, D.~M.; {Lovett}, S.; {Moran}, S.; and {Zhang}, J.
\newblock 2017.
\newblock Active classification with comparison queries.
\newblock In {\em FOCS}.

\bibitem[\protect\citeauthoryear{Kingma and Ba}{2015}]{Kingma2015Adam}
Kingma, D.~P., and Ba, J.
\newblock 2015.
\newblock Adam: {A} method for stochastic optimization.
\newblock In {\em 3rd International Conference on Learning Representations}.

\bibitem[\protect\citeauthoryear{Konyushkova, Sznitman, and
  Fua}{2017}]{Ksenia2017LAL}
Konyushkova, K.; Sznitman, R.; and Fua, P.
\newblock 2017.
\newblock Learning active learning from data.
\newblock {\em Advances in Neural Information Processing Systems 30}.

\bibitem[\protect\citeauthoryear{Lakshminarayanan, Pritzel, and
  Blundell}{2017}]{Balaji2017NIPSUncertaintyEnsembles}
Lakshminarayanan, B.; Pritzel, A.; and Blundell, C.
\newblock 2017.
\newblock Simple and scalable predictive uncertainty estimation using deep
  ensembles.
\newblock {\em Advances in Neural Information Processing Systems 30}.

\bibitem[\protect\citeauthoryear{LeCun and
  Cortes}{2010}]{lecun2010MnistDatabase}
LeCun, Y., and Cortes, C.
\newblock 2010.
\newblock {MNIST} handwritten digit database.
\newblock {\em Proceedings of the IEEE} 86(11):2278--2324.

\bibitem[\protect\citeauthoryear{Lecun \bgroup et al\mbox.\egroup
  }{1998}]{Lecun1998Lenet5}
Lecun, Y.; Bottou, L.; Bengio, Y.; and Haffner, P.
\newblock 1998.
\newblock Gradient-based learning applied to document recognition.
\newblock In {\em Proceedings of the IEEE}.

\bibitem[\protect\citeauthoryear{Lowe}{2004}]{David2004SIFT}
Lowe, D.~G.
\newblock 2004.
\newblock Distinctive image features from scale-invariant keypoints.
\newblock {\em Int. J. Comput. Vision}.

\bibitem[\protect\citeauthoryear{Pang \bgroup et al\mbox.\egroup
  }{2018}]{Pang2018ICMLMetaActiveLearningbyDeepReinforcement}
Pang, K.; Dong, M.; Wu, Y.; and Hospedales, T.~M.
\newblock 2018.
\newblock {Meta-Learning Transferable Active Learning Policies by Deep
  Reinforcement Learning}.
\newblock {\em ArXiv}.

\bibitem[\protect\citeauthoryear{Pop and Fulop}{2018}]{Pop2018DEBAL}
Pop, R., and Fulop, P.
\newblock 2018.
\newblock Deep ensemble bayesian active learning : Addressing the model
  collapse issue in monte carlo dropout via ensemble.
\newblock {\em ArXiv}.

\bibitem[\protect\citeauthoryear{Rosales, Krishnamurthy, and
  Bharat~Rao}{2008}]{Romer2009QBC1}
Rosales, R.; Krishnamurthy, P.; and Bharat~Rao, R.
\newblock 2008.
\newblock Semi-supervised active learning for modeling medical concepts from
  free text.
\newblock {\em In: Proceedings of the Sixth International Conference on Machine
  Learning and Applications}.

\bibitem[\protect\citeauthoryear{Sener and
  Savarese}{2018}]{sener2018ICLRCoreset}
Sener, O., and Savarese, S.
\newblock 2018.
\newblock Active learning for convolutional neural networks: a core-set
  approach.
\newblock In {\em International Conference on Learning Representations}.

\bibitem[\protect\citeauthoryear{Shannon}{2001}]{Shannon2001Entropy}
Shannon, C.~E.
\newblock 2001.
\newblock A mathematical theory of communication.
\newblock {\em SIGMOBILE Mob. Comput. Commun. Rev.}

\bibitem[\protect\citeauthoryear{Su \bgroup et al\mbox.\egroup
  }{2019}]{Su_2019_CVPR_Workshops}
Su, J.~C.; Tsai, Y.~H.; Sohn, K.; Liu, B.; Maji, S.; and Chandraker, M.
\newblock 2019.
\newblock Active adversarial domain adaptation.
\newblock In {\em CVPR Workshops}.

\bibitem[\protect\citeauthoryear{{Tang} \bgroup et al\mbox.\egroup
  }{2017}]{Tang2017ADL}
{Tang}, B.; {Xu}, J.; {He}, H.; and {Man}, H.
\newblock 2017.
\newblock {ADL}: Active dictionary learning for sparse representation.
\newblock In {\em IJCNN}.

\bibitem[\protect\citeauthoryear{{Wang} \bgroup et al\mbox.\egroup
  }{2017}]{Wang2017USALDCSS}
{Wang}, G.; {Hwang}, J.; {Rose}, C.; and {Wallace}, F.
\newblock 2017.
\newblock Uncertainty sampling based active learning with diversity constraint
  by sparse selection.
\newblock In {\em MMSP}.

\bibitem[\protect\citeauthoryear{Wang \bgroup et al\mbox.\egroup
  }{2019}]{wang2019datasetdistillation}
Wang, T.; Zhu, J.-Y.; Torralba, A.; and Efros, A.~A.
\newblock 2019.
\newblock Dataset distillation.
\newblock {\em ArXiv}.

\bibitem[\protect\citeauthoryear{Zhang and
  Chaudhuri}{2014}]{Zhang2014BDisagreementAL}
Zhang, C., and Chaudhuri, K.
\newblock 2014.
\newblock Beyond disagreement-based agnostic active learning.
\newblock {\em ArXiv}.

\bibitem[\protect\citeauthoryear{Zhou and Sun}{2014}]{Jin2014IMSAL}
Zhou, J., and Sun, S.
\newblock 2014.
\newblock Improved margin sampling for active learning.
\newblock In {\em CCPR}.

\end{thebibliography}
\end{document}